\documentclass[article]{acmart}
\usepackage{array} 
\usepackage{booktabs} 
\usepackage{diagbox} 
\AtBeginDocument{%
  }

\setcopyright{acmlicensed}
\copyrightyear{2024}
\acmYear{2024}
\acmDOI{XXXXXXX.XXXXXXX}

\acmJournal{CSUR}
\acmVolume{37}
\acmNumber{4}
\acmArticle{111}
\acmMonth{8}

\begin{document}

\title{Bridging the Gap: Representation Spaces in Neuro-Symbolic AI}

\author{XIN ZHANG}
\orcid{0000-0003-4960-174X}
\affiliation{%
	\institution{Department of Computer Science, Texas Tech University}
	\streetaddress{2500 Broadway}
	\city{Lubbock}
	\state{Texas}
	\postcode{79409}
	\country{USA}}
\email{e-mail: zha19053@ttu.edu}

\author{VICTOR S.SHENG}
\orcid{0000-0003-4960-174X}
\affiliation{%
	\institution{Department of Computer Science, Texas Tech University}
	\streetaddress{2500 Broadway}
	\city{Lubbock}
	\state{Texas}
	\postcode{79409}
	\country{USA}}
\email{e-mail: victor.sheng@ttu.edu}

\renewcommand{\shortauthors}{Trovato et al.}

\begin{abstract}
Neuro-symbolic AI is an effective method for improving the overall performance of AI models by combining the advantages of neural networks and symbolic learning. However, there are differences between the two in terms of how they process data, primarily because they often use different data representation methods, which is often an important factor limiting the overall performance of the two. From this perspective, we analyzed 191 studies from 2013 by constructing a four-level classification framework. The first level defines five types of representation spaces, and the second level focuses on five types of information modalities that the representation space can represent. Then, the third level describes four symbolic logic methods. Finally, the fourth-level categories propose three collaboration strategies between neural networks and symbolic learning. Furthermore, we conducted a detailed analysis of 46 research based on their representation space.
\end{abstract}

\ccsdesc[500]{Do Not Use This Code~Generate the Correct Terms for Your Paper}
\ccsdesc[300]{Do Not Use This Code~Generate the Correct Terms for Your Paper}
\ccsdesc{Do Not Use This Code~Generate the Correct Terms for Your Paper}
\ccsdesc[100]{Do Not Use This Code~Generate the Correct Terms for Your Paper}

\keywords{Do, Not, Us, This, Code, Put, the, Correct, Terms, for,
  Your, Paper}

\received{20 September 2024}
\received[revised]{12 September 2024}
\received[accepted]{5 September 2024}

\maketitle
\section{Introduction}
Neuro-symbolic AI is a promising paradigm that combines both the powerful learning abilities of neural networks and the logical reasoning of symbolic AI to address complex AI problems. However, although the cooperation between these two seems natural, the difference in their representation is obviously not negligible.

Prof. Henry Kautz proposed a taxonomy of Neuro-Symbolic Systems in the AAAI 2020. In addition, many researchers have conducted relevant reviews of the recent neuro-symbolic AI from different perspectives. As Fig.1 shows, \citet{ref196acharya2023neurosymbolic} proposed a new classification method, which classified and discussed the application of existing neuro-symbolic AI by the role of neural and symbolic parts: learning for reasoning, reasoning for Learning, and learning-reasoning. \citet{ref197garcez2015neural} proposed a taxonomy that includes sequential, nested, cooperative, and compiled neuro-symbolic AI based on the six types introduced by Henry Kautz. In addition, some reviews focus on cross-field integration and applications. For example, \citet{ref198berlot2021neuro} reviewed neuro-symbolic VQA (visual question answering) from the perspectives of AGI (artificial general intelligence) desiderata. \citet{ref199marra2024statistical} conducted a comprehensive review on integrating neuro-symbolic and statistical relational artificial intelligence based on seven dimensions. Additionally, \citet{ref200belle2023statistical} explored the integration between SRL (statistical relational Learning) and neuro-symbolic Learning based on distinctions between subjective probabilities and the semantics of random worlds, the significance of infinite domains and random world semantics, and the application of probability to formulas and quantifiers. \citet{ref201kleyko2022survey}\citet{ref202kleyko2023survey} are two-part summaries of HDC (hyperdimensional computing) and VSA (vector symbolic architecture) from known computing models, conversion of various input data types to high-dimensional distributed representations, related applications, cognitive computing and architecture, and directions for future work. \citet{ref203delong2023neurosymbolic,ref204zhang2020neural,ref205singh2023neuro,ref206lamb2020graph,ref207khan2020neuro,ref208zhang2021neural} conducted multi-faceted summaries of graph theory and ontological inference based on neuro-symbolic reasoning. \citet{ref209panchendrarajan2024synergizing} discussed a hybrid method combining machine learning and symbolic methods, focusing on three sub-fields of natural language processing: understanding, generation, and reasoning.

\begin{figure}[h]
	\centering
	\includegraphics[width=\linewidth]{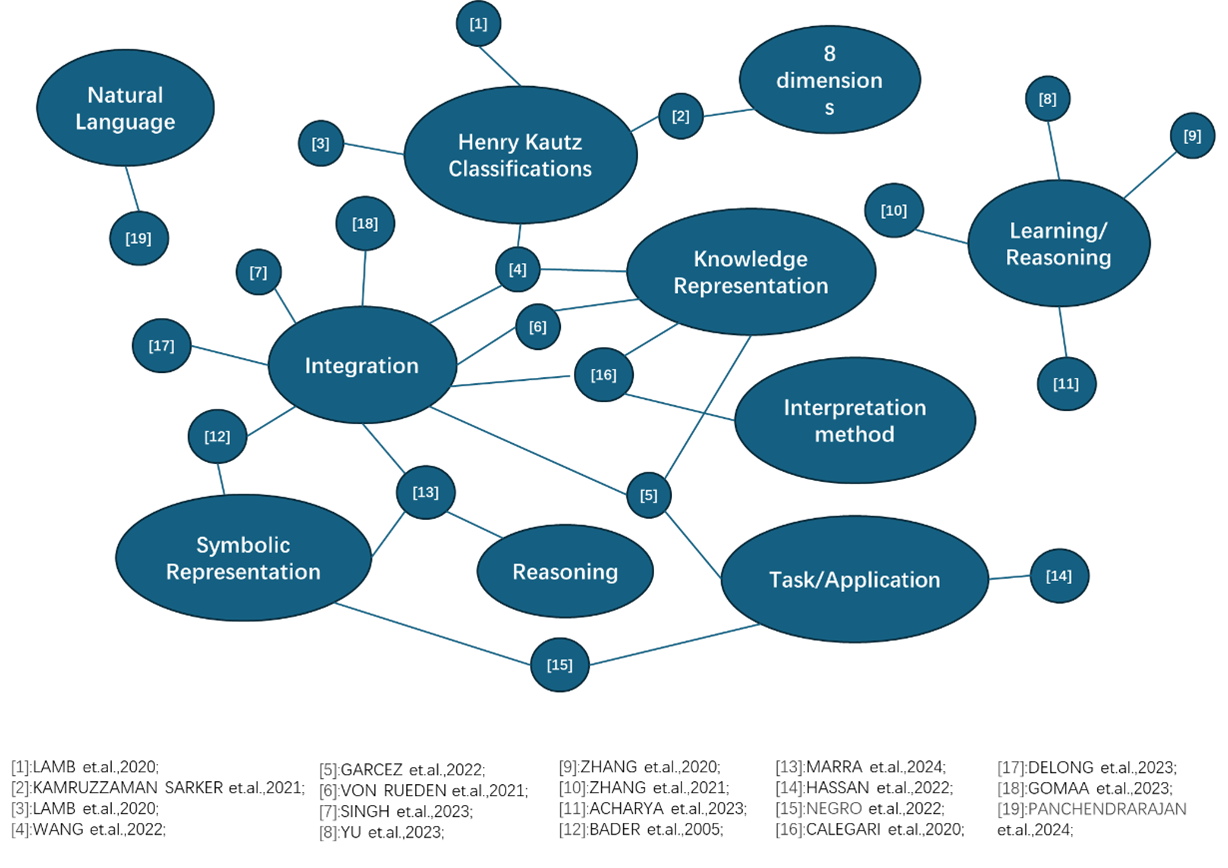}
	\caption{Current reviews on neuro-symbolic AI}
\end{figure}

This survey supplements the existing reviews mentioned above. It also aims to help beginners quickly understand the latest research trends and typical working principles in neuro-symbolic AI from the perspectives of representation space. Furthermore, we focus on the representation capacity of different modalities and its support for the representation of neural networks and symbolic learning.
\section{Types of Neuro-symbolic AI Based on Representation Space}
In this article, modal refers to the modality of input data, so the single-modal model describes a method that can only process one data type. In contrast, the multi-modal model can process more than one data type. In addition, non-heterogeneous and heterogeneous refer to whether the representation space can simultaneously support the embedding vectors of neural networks and symbolic logic instead of representing them in the other's way. A representation space that can only support one is called a non-heterogeneous representation space. Otherwise, it is a heterogeneous representation space. Combining the upper two categorizing methods, we divide existing neuro-symbolic AI research into five types: uni-modal non-heterogeneous, multi-modal non-heterogeneous, single-modal heterogeneous, multi-modal heterogeneous, and dynamic adaptive model.
\begin{table}[ht]
	\centering 
	\caption{Neuro-Symbolic AI's classification by representation space} 
	\label{tab:neuro_symbolic}
	\begin{tabular}{
			>{\centering\arraybackslash}m{3cm} 
			>{\centering\arraybackslash}m{2cm} 
			>{\centering\arraybackslash}m{2cm} 
			>{\centering\arraybackslash}m{3cm} 
			>{\centering\arraybackslash}m{3cm} 
		}
		\toprule 
		& Single Modal Data & Multi-Modal Data & Neural Network OR Symbolic Logic Representation & Neural Network AND Symbolic Logic Representation \\
		\midrule 
		Single-modal and non-heterogeneous & $\checkmark$ & & $\checkmark$ & \\
		Multimodal and non-heterogeneous & & $\checkmark$ & $\checkmark$ & \\
		Single-modal and heterogeneous & $\checkmark$ & & & $\checkmark$ \\
		Multimodal and heterogeneous & & $\checkmark$ & & $\checkmark$ \\
		Dynamic adaptive & $\checkmark$ & $\checkmark$ & $\checkmark$ & $\checkmark$ \\
		\bottomrule
	\end{tabular}
\end{table}

	The table clearly shows the definitions of each category:
\begin{enumerate}
	\item Single modal non-heterogeneous neuro-symbolic AI: 
	neural networks extract features from single-modal data, and the representation space only supports one type of representation.
	\item Multi-modal non-heterogeneous neuro-symbolic AI: neural networks extract features from multi-modal data, and the representation space only supports one of the neural network or logical symbol representation.
	\item Single-modal heterogeneous neuro-symbolic AI: neural networks extract features from single-modal data, and the representation space can support both neural networks and logical symbol representations.
	\item Multi-modal heterogeneous neuro-symbolic AI: neural networks extract features from multi-modal data, and the representation space can support both neural network and logical symbol representation.
	\item Dynamic adaptive neuro-symbolic AI: The representation space can be dynamically adjusted and optimized according to task requirements, that is, to fulfill all the above four classifications dynamically.
\end{enumerate}	
This study investigated 191 existing neuro-symbolic AI studies since 2013, including 175 that used single-modal non-heterogeneous representation methods and 13 research that used multi-modal non-heterogeneous hybrid representations. There are two studies on single-modal heterogeneous representation methods and one on multi-modal heterogeneous models. No studies are currently using multi-modal heterogeneous and dynamic adaptive representation methods.
\begin{table}[ht]
	\centering 
	\caption{Statistics of articles in each category}
	\label{tab:neuro_symbolic} 
	\begin{tabular}{
			>{\centering\arraybackslash}m{6cm} 
			>{\centering\arraybackslash}m{5cm} 
		}
		\toprule 
		Classification by Representation Space & Number of Papers \\
		\midrule 
		Single-modal and non-heterogeneous & 175 \\
		Multimodal and non-heterogeneous & 13 \\
		Single-modal and heterogeneous & 2 \\
		Multimodal and heterogeneous & 1 \\
		Dynamic adaptive & 0 \\
		\bottomrule 
	\end{tabular}
\end{table}
\section{Single-modal Non-heterogeneous Neuro-symbolic AI}
We classify 175 neuro-symbolic AI studies into five sub categories by data type that has been processed: text, image, environment and state, numerical and mathematical expressions, and structured data.
\subsection{Text}
This category covers 51 studies in which neural networks extract features from text data and then process them using logical symbolic methods. Additionally, these studies can be grouped into four divisions based on the type of symbolic logic: logical rules and programming, symbolic representation and structure, knowledge graphs and databases, mathematics, and numerical operations.
\begin{table}[ht]
	\centering 
	\caption{Input data statistics of Neuro-symbolic research} 
	\label{tab:neuro_symbolic}
	\begin{tabular}{
			>{\centering\arraybackslash}m{4cm} 
			>{\centering\arraybackslash}m{2.5cm} 
			>{\centering\arraybackslash}m{2cm} 
			>{\centering\arraybackslash}m{2cm}
			>{\centering\arraybackslash}m{2cm} 
		}
		\toprule
		\diagbox{Symbolic form}{Data Type} & Logic rules and programming & Symbolic representation and structure & Knowledge graphs and databases & Mathematical and numerical operations \\
		\midrule 
		Text & 32 & 6 & 12 & 1 \\
		Image & 35 & 8 & 5 & 3 \\
		Environment and State-Aware Data & 14 & 4 & 0 & 1 \\
		Numerical and Mathematical Expressions & 10 & 2 & 0 & 15 \\
		Structured Data & 15 & 2 & 10 & 0 \\
		\bottomrule
	\end{tabular}
\end{table}

\subsubsection{Symbolic:Logic Rules and Programming}
This portfolio includes 32 studies, all extracting features from text like natural language, programming language, and descriptions of specific fields, then converting features into a form that can be processed by symbolic logic through semantic parsing. This process bridges data-based pattern recognition and rule-based logical reasoning. Research within this combination can be divided into three groups based on how neural networks and symbolic logic cooperate. In the rest of this part of the review, we will default to the classification model in this section for research statistics.
\begin{enumerate}
	\item Neuro-symbolic generation: features are extracted by neural networks, and then these features are transformed into a form that a symbolic logical module can handle. Research in this category includes \cite{ref1Pan2023logic,ref2alon2022neuro,ref3liu2022neural,ref4hooshyar2024temporal,ref5davis2022neurolisp,ref6chaudhury2021neuro,ref7pallagani2022plansformer,ref8karpas2022mrkl,ref9nye2021improving,ref10ashcraft2023neuro,ref11bonzon2017towards,ref12shakya2021student,ref13chen2019neural,ref14qin2021neural,ref15baugh2023neuro,ref16devlin2017semantic,ref17arabshahi2021conversational,ref18zhu2022neural,ref19nunez2023nesig}.
	\item Symbolic-neural enhancement: to enhance neural networks by integrating high-level knowledge, such as symbolic logical rules, knowledge, or structured information provided by symbolic logic for better feature interpretation or learning process. \cite{ref20liang2016neural,ref21kapanipathi2020leveraging,ref22arakelyan2022ns3,ref23chaudhury2023learning,ref24tran2017unsupervised,ref24tran2017unsupervised,ref25liu2023weakly,ref27cosler2024neurosynt} all fall into this category.
	\item Neural-symbolic collaboration: a two-way collaborative learning process. The features extracted by neural networks are converted for symbolic logic, and rules from symbolic logic are fed back to the neural network. Research in this category includes \cite{ref28saha2021weakly,ref29ref3Galassi,ref30zhang2023natural,ref31schon2021negation,ref32ying2023neuro}.
\end{enumerate}	
\begin{table}[ht]
	\centering 
	\caption{Take text as input and logic rules and programming as symbolic method} 
	\label{tab:neuro_symbolic} 
	\begin{tabular}{
			>{\centering\arraybackslash}m{4cm}
			>{\centering\arraybackslash}m{9cm}
		}
		\toprule
		Method of Collaboration & Papers \\
		\midrule
		Neuro-symbolic generation & \cite{ref1Pan2023logic,ref2alon2022neuro,ref3liu2022neural,ref4hooshyar2024temporal,ref5davis2022neurolisp,ref6chaudhury2021neuro,ref7pallagani2022plansformer,ref8karpas2022mrkl,ref9nye2021improving,ref10ashcraft2023neuro,ref11bonzon2017towards,ref12shakya2021student,ref13chen2019neural,ref14qin2021neural,ref15baugh2023neuro,ref16devlin2017semantic,ref17arabshahi2021conversational,ref18zhu2022neural,ref19nunez2023nesig}\\
		Symbolic-neural enhancement & \cite{ref20liang2016neural,ref21kapanipathi2020leveraging,ref22arakelyan2022ns3,ref23chaudhury2023learning,ref24tran2017unsupervised,ref24tran2017unsupervised,ref25liu2023weakly,ref27cosler2024neurosynt} \\
		Neural-symbolic collaboration & \cite{ref28saha2021weakly,ref29ref3Galassi,ref30zhang2023natural,ref31schon2021negation,ref32ying2023neuro} \\
		\bottomrule 
	\end{tabular}
\end{table}

Among these studies, \citet{ref20liang2016neural} proposed an NSM (neural symbolic machine) that combines neural networks and symbolic logic to perform efficient discrete operations on large knowledge bases. NSM uses a neural programmer module to accept natural language input through questions and descriptions, extracts semantics through a sequence-to-sequence model and generates executable programs by mapping semantics to a series of tokens. The manager module provides weak supervision signals in the form of correct answers to tasks, indicating the degree of task completion through rewards. Programmers need to learn from the rewards the manager provides and find the proper program. Finally, NSM uses a Lisp interpreter to perform non-differentiable operations to execute programs generated by the programmer module. To solve the problem of finding the correct program encountered when training from question-answer pairs, NSM prunes the programmer's search space by checking the syntax and semantics of the generated program, i.e., checking whether the generated program will cause syntax or semantic errors And filter out invalid program sequences to improve training efficiency. Symbolic logic exists in program expressions and Lisp interpreters in the above process. The former constructs a program sequence that represents specific operations generated by a neural network and represents a particular operation by converting natural language into code—probabilistic generative models of environments. When trained using only question-answer pairs, NSM achieves new state-of-the-art performance on the WebQuestionSSP dataset without any feature engineering or domain-specific knowledge, demonstrating the power of NSM by integrating the statistical learning capabilities of neural networks and symbolic logic. Reasoning capabilities, which can effectively learn from weakly supervised signals in semantic parsing tasks using large-scale knowledge bases.

\citet{ref1Pan2023logic} proposed LOGIC-LM, a method to solve logic problems by combining a LLM (large language model) with a symbolic solver. This method effectively links natural language processing and deterministic logical reasoning using three stages: problem formulation, symbolic reasoning, and result interpretation. LOGIC-LM first uses LLM to interpret and translate the fundamental entities, facts, and logical rules in the natural language statement of the problem into predicates, variables, and logical expressions in logic. LOGIC-LM then operates on the symbolic representation using a deterministic symbolic solver and derives the answer or solution to the given problem through logical reasoning. At the same time, the deterministic nature of the solver ensures that conclusions are logically consistent and traceable. Finally, LOGIC-LM uses a self-refining module to iteratively improve the accuracy of symbolic translation based on feedback from the symbolic solver. In cases where the initial symbolic formulation leads to errors or is deemed inaccurate, the self-refining module utilizes input from the solver—error message to modify and improve the formula. In the above process, symbolic logic exists in the form of logic programming languages, first-order logic, constraint satisfaction problems, and Boolean satisfiability problems. The effectiveness of LOGIC-LM has been demonstrated on multiple logical reasoning data sets ranging from deductive reasoning to constraint satisfaction problems, indicating that this method provides a feasible idea for solving the limitations of large language models in reliable logical reasoning.

\citet{ref29ref3Galassi} proposed a neural symbolic argument mining framework that improves argument mining performance by combining neural networks and symbolic logic. This method first uses neural networks such as recurrent neural networks, convolutional neural networks, and transformer architectures to extract features from text data such as academic articles, social media content, and legal documents and automatically identify argument components, such as claims, reasons, and evidence in the article. And the relationship between them, such as support or opposition. This study proposes to use PLP (probabilistic logic programming) to fuse neural network output and symbolic logic representation. Specifically, the PLP framework uses logical rules, such as defeasible rules with probabilistic labels attached to represent uncertainty. It uses probabilistic logical rules by taking the argument components and relationships the neural network identifies as input for reasoning and analysis. This method can simultaneously identify argument components and analyze argument relationships in a single learning process. It can achieve global decision-making adjustments by introducing rules and constraints within the training phase. Symbolic logic in this study exists in the form of structured arguments and abstract arguments, where the former represents knowledge by defining a formal language and specifying how to construct arguments and counter-arguments from that knowledge, such as using strict rules and overturn rules to express the structure and content of arguments. Abstract arguments deal with logical inconsistencies by focusing on high-level relationships between arguments. The method proposed in this study can handle complex reasoning tasks more effectively than traditional argument mining.

\subsubsection{Symbolic:Symbolic Representation and Structure}
This category includes six studies. The neural network extracts features from text data by extended short-term memory networks, universal sentence encoder, InferSent sentence embeddings, or Bert models, and then converts text input into structured representations by various methods, such as using a symbolic stack machine to manipulate text sequences or a grammatical structure of sentences like syntactic parse trees or generating symbolic expressions to represent the solution process of mathematical problems. Among them, Research within this group that belongs to symbolic-neural enhancement includes \cite{ref33pinhanez2020using,ref34chen2020compositional,ref35hu2022fix}and those belonging to the symbolic-neural enhancement classification include \cite{ref36chrupala2019correlating,ref37gaur2023reasoning,ref38van2017linking}.
\begin{table}[ht]
	\centering 
	\caption{Take text as input and symbolic representation and structure as symbolic method}
	\label{tab:neuro_symbolic} 
	\begin{tabular}{
			>{\centering\arraybackslash}m{4cm} 
			>{\centering\arraybackslash}m{9cm} 
		}
		\toprule 
		Method of Collaboration & Papers \\
		\midrule 
		Symbolic-neural enhancement & \cite{ref33pinhanez2020using,ref34chen2020compositional,ref35hu2022fix} \\
		Neural-symbolic collaboration & \cite{ref36chrupala2019correlating,ref37gaur2023reasoning,ref38van2017linking} \\
		\bottomrule 
	\end{tabular}
\end{table}

\citet{ref33pinhanez2020using} proposed a method to improve the accuracy of intent recognition by utilizing meta-knowledge embedded in intent recognition identifiers in dialogue systems. From the perspective of existing knowledge, obtaining structured and complete knowledge from text or humans is a challenge. This study provides an efficient approach to knowledge acquisition in Neuro-Symbolic systems by showing how to leverage existing taxonomies of prototypes embedded in intention identifiers. The method first extracts features from user utterances or sentences used for intent recognition in dialogue systems through neural networks and generates sentences by embedding a set of intent identifiers into another continuous vector space to create embedded representation. Then, meta-knowledge is used to map the representation in this vector space to another representation of the intent identifier in the vector representation space embedded through proto taxonomies, that is, by analyzing the developer's implicit prototype taxonomy in the intent identifier to capture high-level knowledge structures and use this structure to improve the model's intent recognition capabilities. Symbolic logic exists in two forms in this process: one is meta-knowledge expressed through prototype taxonomies, and the other is a structure like a knowledge graph composed of these prototype taxonomies. The prototype taxonomy reflects the structured knowledge developers embed in intent identifiers by connecting high-level, symbolic concepts shared between different intents. These knowledge structures are informal but structured, describing relationships and hierarchies between different intentions. Experimental results show that embedding meta-knowledge in this way can improve the accuracy of intent recognition in most cases. Identifying "out-of-scope" samples can significantly improve identification accuracy and reduce the proportion of false acceptance rates. At the same time, the method can automatically mine and utilize the knowledge embedded in the dialogue system without the direct intervention of experts.

\citet{ref34chen2020compositional} proposed a NeSS (Neural-Symbolic Stack Machine) as a machine operation controller by integrating the symbolic stack machine into a sequence-to-sequence generation framework. Specifically, the method uses neural networks to extract features from input sequences in the source language and output sequences in the target language. These text sequences contain commands or instructions to guide the operation of the neural symbolic machine. Then, the neural network is used to act as a controller to generate a series of execution traces as operation instructions based on the characteristics of the input sequence. These instructions are subsequently executed by a symbolic stack machine with sequence operation capabilities. The input sequence is processed through a series of recursive processing and sequence operations, and the target output sequence is generated to realize the combined understanding and transformation of the input sequence. Symbolic logic in NeSS mainly exists in two forms: Symbolic Stack Machine and Operational Equivalence. The former is the core component of NeSS. It supports recursive and sequence operations through symbolic operations, such as stack push, stack pop, sequence generation, and other instructions to realize the combined processing of input sequences and the generation of output sequences. At the same time, the symbolic stack machine supports recursion so that the entire sequence can be broken down into components and processed separately. Operational equivalence is a crucial concept used by NeSS to improve generalization capabilities. It identifies and classifies semantically similar components by comparing the similarity of execution traces generated by different input sequences, further promoting the model to learn the rules for combining components. Experimental shows that NeSS performs well in four benchmark tests that require combinatorial generalization, including the benchmark test of SCAN language-driven navigation task, the combined instruction task of few-shot learning, the combined machine translation benchmark test, and the context-free grammar parsing task. Achieving 100\% generalization performance shows that NeSS can understand and generate sequences that comply with given rules and generalize the learned knowledge to new, unseen combinations.

In addition, in \cite{ref36chrupala2019correlating}, the symbolic output is used to verify the correctness of solving mathematical problems. In contrast, the method proposed by \cite{ref37gaur2023reasoning} uses symbolic output to explain the structure or semantics of the sentence. The method proposed by \cite{ref38van2017linking} emphasizes the process of automatically generating and utilizing symbols from sensory data, that is, using an incremental learning process to extract structures and processes from input data and generate symbols bottom-up, where each symbol represents A pattern or concept in the input data. In addition, the method uses working memory to bind relationships between symbols and control structures, simulating how the human brain works when processing complex conceptual structures. The above examples illustrate how to effectively combine continuous vector space representation and high-level, discrete, structured knowledge representation, how to combine the learning capabilities of neural networks with symbolism, and how the precise rules and structures of logic are used to improve the understanding, reasoning, generalization, and explanation capabilities of the model.
\subsubsection{Symbolic:Knowledge Graphs and Databases}
This category includes 12 studies in which neural networks extract features from text, and symbolic logic exists in knowledge graphs, first-order logic facts, and ontologies, representing explicit rules, entities, and relations among entities to support reasoning and decision-making. Research within this group that belongs to neural-symbol generation includes \cite{ref39kimura2021neuro,ref40verga2020facts,ref41baran2022linguistic,ref42hwang2021comet,ref43verga2021adaptable,ref44bosselut2021dynamic}. Those belonging to the symbolic-neural enhancement classification include \cite{ref46jain2023reonto}. While \cite{ref47tong2023neolaf,ref48cingillioglu2021pix2rule,ref49ma2019towards,ref50hu2022empowering,ref51hu2023chatdb}belong to Neuro-Symbolic collaboration.
\begin{table}[ht]
	\centering 
	\caption{Take text as input and knowledge graphs and databases as symbolic method}
	\label{tab:neuro_symbolic} 
	\begin{tabular}{
			>{\centering\arraybackslash}m{4cm} 
			>{\centering\arraybackslash}m{9cm} 
		}
		\toprule 
		Method of Collaboration & Papers \\
		\midrule 
		Neuro-symbolic generation & \cite{ref39kimura2021neuro,ref40verga2020facts,ref41baran2022linguistic,ref42hwang2021comet,ref43verga2021adaptable,ref44bosselut2021dynamic} \\
		Symbolic-neural enhancement & \cite{ref46jain2023reonto}\\
		Neural-symbolic collaboration & \cite{ref47tong2023neolaf,ref48cingillioglu2021pix2rule,ref49ma2019towards,ref50hu2022empowering,ref51hu2023chatdb} \\
		\bottomrule
	\end{tabular}
\end{table}

\citet{ref40verga2020facts} proposed a method to improve the model's performance on knowledge-intensive tasks by helping the neural network model learn from large-scale text data and directly interact with the structured knowledge base. The neural network part of the method is based on large-scale pre-trained language models to learn syntax, semantics, and other features from large amounts of text data and understand the meaning of words, phrases, and sentences by capturing the nuances of language. Then, the context embedding representation generated by a large pre-trained language model is used as a query to retrieve triple information related to the current context in the knowledge base. The retrieval results are converted back into a form understandable by the neural network and used together with the contextual embedding of the text for the final task, such as answering a question. In this approach, symbolic logic exists as triples in an external knowledge base. Through an explicit interface, factual information in symbolic logic is combined with the underlying knowledge encoded in the neural network. This study, therefore, explicitly integrates symbolic logic, such as facts from an external knowledge base, into a large pre-trained language model so that part of the latter's decisions can be understood and explained by directly looking at the facts used. In addition, this approach allows the model's behavior to be changed by simply updating facts in the knowledge base, avoiding the cost of retraining the model.

\citet{ref46jain2023reonto} proposed ReOnto (Relation Extraction Ontology), which combines graph neural networks and publicly accessible ontology as prior knowledge to complete relationship extraction in biomedical texts by identifying the sentence relationships between two entities. This method learns feature representations of entity pairs by encoding the entities in the sentence and the underlying relationships between them. That is, entity pairs are embedded into a graph structure, where entities serve as nodes and potential relationships serve as edges, and then the complex interactions between entities are captured through graph neural networks. Then, the encoded path information and the sentence representation processed by the graph neural network are combined by calculating the semantic similarity between the paths extracted from the ontology and the entity relationships in the sentences to predict the relationships between entity pairs jointly. Symbolic logic in ReOnto exists as relationship paths in the ontology, representing the paths connecting two entities through a series of logical relationships. Specifically, this method first finds the direct relationship path between two entities by querying ontology. If the direct path does not exist, it uses ontology reasoning to find the indirect multi-hop relationship path connecting the two entities. In addition, the relational path also includes expressive axioms in the ontology, such as logical quantifiers that include existential quantifiers $\exists$, universal quantifiers $\forall$, and set operations, including union and intersection, to enrich and expand the meaning of the relational path. Experimental results show that the ReOnto method outperforms all baseline methods on two public biomedical datasets (BioRel and ADE), improving by approximately 3\%. This result demonstrates the effectiveness of ReOnto in biomedical relationship extraction tasks.

\citet{ref50hu2022empowering} proposed a model architecture OREOLM (knOwledge REasOning empowered Language Model) that improves the performance of open domain question answering by integrating knowledge graph reasoning of symbolic logic and neural networks. The core of this method is to enable the language model to work together with a differentiable knowledge graph reasoning module through the Knowledge Interaction Layers embedded in the language model. OREOLM uses a transformer-based language model to extract features from natural language text by identifying critical entities in the question and its context and generating queries or relationship predictions related to these entities. The transformation mechanism of the neural network and symbolic logic in this method is implemented by the knowledge interaction layer inserted between the transformer layers. Specifically, for each essential entity identified in the question, the language model is first based on the context of each question and the language model's understanding of possible relationships between entities to predict a distribution of relationships associated with those entities. This distribution is then used to guide the knowledge graph reasoning module with actual instructions for further traversing the graph along the predicted relationships. Next, the knowledge graph reasoning module performs a contextualized random walk based on the instructions provided by the language model and collects and summarizes information along the predicted path. This collected information is then encoded into embedding vectors and integrated into the language model, which further serves as additional contextual information to help the language model understand the question and generate answers. Experiments show that OREOLM achieves significant performance improvements on several benchmark datasets of Open Domain Question Answering, especially in closed-book settings, especially when dealing with complex problems requiring multi-hop or missing relationship inference. Significantly. This shows that OREOLM can improve answers to existing facts and discover new knowledge through reasoning.
\subsubsection{Symbolic:Mathematical and Numerical Operations}
This portfolio includes a total of one study. \citet{ref52flach2023neural} focuses on using $\lambda$-calculus for encoding and calculation and utilizing logical symbols for calculation by learning to perform reductions in $\lambda$-calculus. This research includes detailed hypotheses (H1 and H2) regarding the transformer model's capabilities: H1 asserts that the Transformer can learn to perform a one-step computation in $\lambda$-calculus. At the same time, H2 proposes that it can execute complete computations. Specifically, This method uses the Transformer model to extract features from the $\lambda$-terms in text form generated by using the grammatical rules of the $\lambda$ calculus. The output is the new $\lambda$-terms after the $\beta$-reduction of these terms; the free variables in the function body are replaced with actual parameters. The $\lambda$ calculus includes the abstract definition and application of functions. It is a formal system used to express function abstraction and function application. It is the theoretical basis of functional programming languages and Turing Complete and can theoretically represent any computable problem. This model can support the learning and research of functional programming languages and simplify expressions through $\lambda$ calculus rules to build more competent code editors and compilers. The transformer model shows high accuracy in performing single-step and multi-step beta-reduction tasks. The model achieved a maximum accuracy of 99.73\% for the One-Step Beta Reduction task. In the Multi-Step Beta Reduction task, the model's accuracy is as high as 97.70\%. Even when the output is not entirely predicted correctly, the string similarity index usually exceeds 99\% , showing that the transformer model can effectively learn and perform computational tasks based on $\lambda$ calculus.
\subsection{Image}
This category includes 51 research studies, all extracting low-level features from image data by neural networks and then using symbolic logic for high-level reasoning and decision-making. These studies involve four sub-categories of logical symbolic methods: logical rules and programming, symbolic representation and structure, knowledge graphs and databases, and mathematics and numerical operations.
\subsubsection{Symbolic:Logic Rules and Programming}
This portfolio includes a total of 35 studies in which neural networks extract features such as objects, the structure of scenes, or other perceptual information from images or visual data and then apply logical rules, predicate logic, and probabilistic logic programming to process features for further understanding, inferring, and decision-making. This combination includes basic applications such as primary image classification and handwritten formula evaluation, as well as higher-level decision-making and reasoning tasks, such as visual relationship detection and abstract logical reasoning, which show that the combined method has great potential in multiple fields and tasks. Among these studies, those belonging to the neural-symbol generation classification include \cite{ref53shindo2021neuro,ref54tsamoura2021neural,ref55cunnington2023ffnsl,ref56feinman2020generating,ref57mao2019neuro,ref58lyu2019sdrl,ref59garnelo2016towards,ref60amizadeh2020neuro,ref61apriceno2021neuro,ref62hsu2023ns3d,ref63alford2021neurosymbolic,ref64xie2022neuro,ref65dang2020plans,ref66cheng2023transition,ref67zhu2023tgr,ref68gopinath2018symbolic,ref69stehr2022probabilistic,ref70yu2022probabilistic,ref71li2024softened,ref72thomas2023neuro,ref73le2021scalable,ref74fadja2022neural,ref76gal2015bayesian,ref77cingillioglu2022end,ref78garcez2018towards,ref79li2020closed,ref80aspis2022embed2sym} ; research belonging to the symbolic-neural enhancement classification including \cite{ref81dragone2021neuro}; studies belonging to neural-symbolic collaboration include \cite{ref82manigrasso2023fuzzy,ref83cunnington2022neuro,ref84van2023nesi,ref85manhaeve2019deepproblog,ref86ahmed2023semantic,ref87bennetot2019towards,ref194wang2023rapid}.
\begin{table}[ht]
	\centering 
	\caption{Take image as input and logic rules and programming as symbolic method} 
	\label{tab:neuro_symbolic} 
	\begin{tabular}{
			>{\centering\arraybackslash}m{4cm} 
			>{\centering\arraybackslash}m{9cm} 
		}
		\toprule 
		Method of Collaboration & Papers \\
		\midrule 
		Neuro-symbolic generation & \cite{ref53shindo2021neuro,ref54tsamoura2021neural,ref55cunnington2023ffnsl,ref56feinman2020generating,ref57mao2019neuro,ref58lyu2019sdrl,ref59garnelo2016towards,ref60amizadeh2020neuro,ref61apriceno2021neuro,ref62hsu2023ns3d,ref63alford2021neurosymbolic,ref64xie2022neuro,ref65dang2020plans,ref66cheng2023transition,ref67zhu2023tgr,ref68gopinath2018symbolic,ref69stehr2022probabilistic,ref70yu2022probabilistic,ref71li2024softened,ref72thomas2023neuro,ref73le2021scalable,ref74fadja2022neural,ref76gal2015bayesian,ref77cingillioglu2022end,ref78garcez2018towards,ref79li2020closed,ref80aspis2022embed2sym} \\
		Symbolic-neural enhancement & \cite{ref81dragone2021neuro} \\
		Neural-symbolic collaboration & \cite{ref82manigrasso2023fuzzy,ref83cunnington2022neuro,ref84van2023nesi,ref85manhaeve2019deepproblog,ref86ahmed2023semantic,ref87bennetot2019towards,ref194wang2023rapid} \\
		\bottomrule 
	\end{tabular}
\end{table}

Among them, \citet{ref71li2024softened} proposed a neural symbolic learning framework to solve the bridging problem between neural network training and symbolic constraint solving. This framework avoids the time-consuming state space search process by introducing a softened symbolic grounding process, optimizing the Boltzmann distribution of symbolic solutions, and adopting an annealing mechanism. The method can extract features from images, such as handwritten arithmetic expressions and visual Sudoku, and identify patterns and structures by learning deep representations of the input data. The research then enabled the conversion between neural networks and symbolic logic through "softened symbol grounding." It then maps the features identified and extracted by the neural network to the potential symbolic space, such as recognized numbers, operators, etc., using the Boltzmann distribution model and MCMC sampling technology to bridge the differences between the continuous feature space of the neural network and the discrete decision space of symbolic logic. Then, the input is fed into a symbolic logic system to generate output. In this method, the symbolic logic part exists as predefined symbolic constraints or rules. These symbolic constraints represent the logical structure and rules of the problem, such as the evaluation rules of arithmetic expressions, Sudoku problem-solving rules, etc., for Neural networks that provide a structured reasoning framework. Experimental results show that this research performs better than existing methods on multiple neural symbols learning tasks such as handwritten formula evaluation, visual Sudoku classification, and shortest path prediction of weighted graphs.

\citet{ref53shindo2021neuro} proposed a NSFR (Neuro-Symbolic Forward Reasoner), differentiable forward chaining based on first-order logic to optimize deriving new facts from known facts and rules through optimization algorithms such as gradient descent. The neural network in this method extracts features from visual data and directly maps the extracted object representation through the object attributes, such as color and shape, output by the neural network to facts as atoms in symbolic logic, and then uses this probabilistic symbolic representation of ground atomic forms for logical reasoning. NSFR approximates logical operations through differentiable forward-chain reasoning, unlike traditional symbolic logic reasoning. This process can be performed within the gradient descent framework and optimized through backpropagation. In NSFR, symbolic logic mainly defines the relationships between objects and rules for reasoning in first-order logic, allowing the model to understand and process high-level concepts and patterns. Through experiments on the Kandinsky pattern in 2D and the CLEVR-Hans dataset in 3D, NSFR shows its power in understanding and reasoning about complex patterns involving object properties, such as color and shape, and spatial relationships, such as "close" and "up." The upper result means that NSFR can handle tasks that require identifying objects and their attributes in images and high-level reasoning based on this information.

\citet{ref78garcez2018towards} proposed a new method, SRL+CS (Symbolic Reinforcement Learning with Common Sense), that can improve the generalization ability, transfer learning ability, abstraction ability, and interpretability of reinforcement learning. This method introduces the concept of symbolic logic into the standard deep reinforcement learning framework. The method mainly uses convolutional neural networks to process image data and map the visual patterns and structures in the image into abstract symbolic representations. The recognized objects in the image are marked with specific symbols, and the relative positions between them are calculated. Finally, Q-learning is performed based on the state space represented by these symbols, with the goal of learning which action to take in a given state can maximize the future cumulative reward. The symbolically represented state space provides the basis for final decision-making. Inspired by the principle of human common sense, SRL+CS introduces two critical improvements in the learning and decision-making process: updating the Q-value only when the object's state interacts with the agent changes, considering the relative position to objects when making decisions, and giving higher importance to closer ones. Experiments have proven that this research can achieve knowledge transfer and generalization in different environment configurations, especially when testing from a deterministic training environment to a random environment, demonstrating near-perfect zero-shot learning capabilities.

\subsubsection{Symbolic:Symbolic Representation and Structure}
The category includes eight studies where neural networks are responsible for processing continuous, high-dimensional visual inputs, and symbolic logic uses this information or patterns for reasoning or decision-making by mapping extracted features to a set of predefined symbols or concepts. Among these studies, those belonging to the neural-symbol generation classification include \cite{ref88agarwal2021end,ref89asai2020learning,ref90stammer2021right,ref91su2022probabilistic,ref92sarkar2015early,ref93khan2023neusyre}; studies belonging to the neural-symbol collaboration include \cite{ref94feinman2020learning,ref95daniele2022deep}.
\begin{table}[ht]
	\centering 
	\caption{Take image as input and symbolic representation and structure as symbolic method} 
	\label{tab:neuro_symbolic} 
	\begin{tabular}{
			>{\centering\arraybackslash}m{4cm} 
			>{\centering\arraybackslash}m{9cm} 
		}
		\toprule 
		Method of Collaboration & Papers \\
		\midrule 
		Neuro-symbolic generation & \cite{ref88agarwal2021end,ref89asai2020learning,ref90stammer2021right,ref91su2022probabilistic,ref92sarkar2015early,ref93khan2023neusyre} \\
		Neural-symbolic collaboration & \cite{ref94feinman2020learning,ref95daniele2022deep} \\
		\bottomrule 
	\end{tabular}
\end{table}

\citet{ref91su2022probabilistic} proposed a model that utilizes neural networks to extract and learn high-dimensional features from visual data while using symbolic logic to interpret these features within a structured, rule-based framework. The method first learns and extracts high-dimensional features from raw visual data such as handwritten characters, object images, or any visual scene, and encodes different objects, shapes, colors, and sizes into high-dimensional vectors and captures statistical properties and patterns within the image. After that, the continuous feature space is mapped to the discrete symbolic space using methods such as the discretization of feature vectors and the application of logical rules for symbolic reasoning based on learned features, and the results of these symbolic logical operations are converted into corresponding image outputs or decision-making. In the method in this study, symbolic logic mainly exists in the form of structured representations, such as using symbolic image renderers, probabilistic program control processes, and symbolic stroke primitives so that the logic and structure behind the image data can be described and reasoned more clearly. In addition, this method can explicitly integrate expert knowledge or predefined logical rules into the learning and reasoning process through posterior constraints, ensuring that the generated symbolic structure and reasoning output are consistent with human understanding and expectations. Compared with traditional data-driven deep learning models, the model proposed in this study can better capture and understand abstract relationships and concepts in images and has the potential for cross-domain knowledge transfer and application.

\citet{ref92sarkar2015early} proposed a neural symbolic framework for detecting instability in combustion conditions crucial for engine health monitoring and prediction by analyzing many serialized high-speed combustion flame images. This method first extracts the low-dimensional semantic features of the image hierarchically through CNN (Convolutional Neural Network) and identifies the coherent structure in the flame. Then, the structures in feature maps in each image frame are composed of time series to form time series data based on image features. Next, the method uses symbolic time series analysis to convert these time series data into symbolic sequences using symbolic methods, such as maximum entropy partitioning, and then builds a generalized D-Markov machine model and uses state splitting and processes such as merging form a state transition matrix that can describe the transition of a flame from a stable to an unstable state. This matrix captures the dynamic behavior of the flame shape over time and provides a basis for early instability detection. This method can capture precursors on low time scales before the flame shape transitions from stable to unstable. It was verified through a large amount of experimental data collected on swirling flow-stabilized burners under different operating conditions. It was found that it is consistent with the traditional PCA method. In comparison, this method can capture subtle changes in the combustion process, detect thermo-acoustic instability, and apply to different types of combustion systems and conditions, with certain versatility and transfer capabilities.
\subsubsection{Symbolic:Knowledge Graphs and Databases}
The portfolio includes a total of five studies that use neural networks to extract features from visual modalities and then use logical symbolic forms such as knowledge graphs, background knowledge, first-order logic programming, and ontologies to represent and process high-level, regularized knowledge to help the model understand and reason about complex relationships and rules in the field. Among these studies, those belonging to the neural-symbol generation classification include \cite{ref96han2021unifying,ref97wu2022zeroc}; the research \cite{ref98mitchener2022detect}belongs to the symbol-neural enhancement classification; the research belonging to the neural-symbol collaboration includes \cite{ref99tao2024deciphering,ref100diaz2022explainable}.
\begin{table}[ht]
	\centering 
	\caption{Take image as input and Knowledge Graphs and Databases as symbolic method}
	\label{tab:neuro_symbolic} 
	\begin{tabular}{
			>{\centering\arraybackslash}m{4cm} 
			>{\centering\arraybackslash}m{9cm} 
		}
		\toprule 
		Method of Collaboration & Papers \\
		\midrule 
		Neuro-symbolic generation & \cite{ref96han2021unifying,ref97wu2022zeroc} \\
		Symbolic-neural enhancement & \cite{ref98mitchener2022detect} \\
		Neural-symbolic collaboration & \cite{ref99tao2024deciphering,ref100diaz2022explainable} \\
		\bottomrule 
	\end{tabular}
\end{table}

\citet{ref100diaz2022explainable} proposed a X-NeSyL (eXplainable Neural-symbolic learning). This method combines deep learning and symbolic logic and uses knowledge graphs as expert knowledge to improve the performance and interpretability of the model. This process uses a combined convolutional neural network, EXPLANet, to extract high-level visual features from image data and map them to symbolic logic defined in the knowledge graph. Then, it compares the model's predicted output through a training process called SHAP-Backprop and the expected symbolic relationship in the knowledge graph, and feedback on the symbolic logic based on the knowledge graph into the training of the neural network model to ensure that the features and predictions learned by the model are consistent with the knowledge of domain experts. X-NeSyL uses SHapley Additive exPlanations values to quantify the contribution of each identified part to the final classification decision and use this to adjust the final output of the model.

Meanwhile, this interpretability metric, SHAP GED, evaluates a model's interpretability by comparing the degree of alignment between the neural symbolic representation generated by the model and the knowledge graph representation. Experimental results show that the EXPLANet model outperforms baseline models, including MonuNet and the pure ResNet-101 classifier, on the MonuMAI dataset, which shows that combining the knowledge of domain experts can effectively improve the performance of deep learning models on specific tasks. In addition, the experimental results also demonstrate that the linear instance-level weighting scheme improves model interpretability while maintaining good classification performance.
\subsubsection{Symbolic:Mathematical and Numerical Operations}
The portfolio includes three studies in which neural networks extract complex patterns and structures from images, time series, or videos. These models then use symbolic regression to discover the mathematical laws behind the data or probabilistic graphical models to model cause and effect in the data relation. Among these studies, those belonging to the neural-symbol generation classification include \cite{ref101kim2020integration}; those belonging to the neural-symbol collaborative classification include \cite{ref102sansone2023learning,ref103fire2016learning}.
\begin{table}[ht]
	\centering 
	\caption{Take image as input and mathematical and numerical operations as symbolic method} 
	\label{tab:neuro_symbolic} 
	\begin{tabular}{
			>{\centering\arraybackslash}m{4cm} 
			>{\centering\arraybackslash}m{9cm} 
		}
		\toprule 
		Method of Collaboration & Papers \\
		\midrule 
		Neuro-symbolic generation & \cite{ref101kim2020integration} \\
		Neural-symbolic collaboration & \cite{ref102sansone2023learning,ref103fire2016learning} \\
		\bottomrule 
	\end{tabular}
\end{table}

\citet{ref101kim2020integration} proposed an EQL (EQuation Learner) that combines neural networks and symbolic regression. This study allows the entire system to be trained end-to-end through the backpropagation algorithm, making the whole model highly interpretable. First, EQL uses a convolutional neural network to extract and identify digital information in handwritten digit images in the MNIST dataset and performs dynamic system analysis by processing sequence data where the position and speed of moving objects change over time, mining movement characteristics from time series. Then, this method converts implicit, continuous features into explicit, interpretable mathematical equations through symbolic regression or converts the continuous neural network feature space into discrete, symbolic mathematical expressions. On the MNIST arithmetic task, the EQL network could extract numbers from images and successfully learn the addition operation. The EQL network extracted unknown parameters regarding dynamic system prediction from the data. It used these parameters to predict the dynamic system's future state. It proved the EQL network's ability to process and understand dynamic systems and improve model interpretability, promoting scientific discovery and technological innovation.
\subsection{Environment and Situation Awareness Data}
This category includes 19 research results, all of which use neural networks to extract features from visual images, sensor data, environmental status information, etc., and then use symbolic logic, such as logical rules, defining goals and constraints, and expressing high-level knowledge of tasks, to perform rule-based reasoning and decision-making. These studies include four categories of logical symbolic methods: logical rules and programming, symbolic representation and structure, knowledge graphs and databases, and mathematics and numerical operations.
\subsubsection{Symbolic:Logic Rules and Programming}
This portfolio includes 14 studies in which neural networks automatically extract complex features from raw data and then use logical rules, first-order logic formulas, and symbolic action models to express and process structured knowledge to guide neural networks. The network's learning process provides an interpretable decision-making basis for performing precise and complex logical reasoning. Among these studies, those belonging to the neural-symbol generation classification include \cite{ref104yan2023point,ref105akintunde2020verifying}; those belonging to the symbol-neural enhancement classification include \cite{ref106lyu2022knowledge,ref107illanes2020symbolic}; those belonging to the neuro- Research on neural-symbol collaboration includes \cite{ref108yang2018peorl,ref109silver2022learning,ref110sharifi2023towards,ref111anderson2020neurosymbolic,ref112moon2021plugin,ref113chitnis2022learning,ref114singireddy2023automaton,ref115besold2017reasoning,ref116jiang2024multi,ref117hazra2023deep}.
\begin{table}[ht]
	\centering 
	\caption{Take environment and situation awareness data as input and logic rules and programming as symbolic method}
	\label{tab:neuro_symbolic} 
	\begin{tabular}{
			>{\centering\arraybackslash}m{4cm} 
			>{\centering\arraybackslash}m{9cm} 
		}
		\toprule 
		Method of Collaboration & Papers \\
		\midrule 
		Neuro-symbolic generation & \cite{ref104yan2023point,ref105akintunde2020verifying} \\
		symbol-neural enhancement & \cite{ref106lyu2022knowledge,ref107illanes2020symbolic} \\
		Neural-symbolic collaboration & \cite{ref108yang2018peorl,ref109silver2022learning,ref110sharifi2023towards,ref111anderson2020neurosymbolic,ref112moon2021plugin,ref113chitnis2022learning,ref114singireddy2023automaton,ref115besold2017reasoning,ref116jiang2024multi,ref117hazra2023deep} \\
		\bottomrule 
	\end{tabular}
\end{table}

\citet{ref117hazra2023deep} proposed a DERRL (Deep Explainable Relational Reinforcement Learning), to express strategies through logical rules generated by symbolic logic, thus providing explainability for the generation process of each decision or action. This method uses neural networks to extract features expressed in logical forms of relationships and objects from environmental states. It uses these logical representations to learn the environment's dynamic laws and strategies' rules. For example, in the Blocks World game, DERRL uses logical predicates such as top(X) and on(X,Y) to describe the relationship between blocks and express the status of block stacking. The neural network's output is a series of parameters of action rules, which correspond to the logical rules of action decision-making. As in the Blocks World example, the neural network output represents the rules for when and how to move blocks. Next, the rules generated by the neural network satisfy the preset logical constraints by defining a semantic loss function. This process can integrate human prior knowledge into the learning process through axioms. Experiments on multiple environments, such as Countdown Game, Blocks World, Gridworld, etc., show that compared with traditional methods and the latest neural logic reinforcement learning method, DERRL performs better regarding computational efficiency, policy accuracy, and semantic constraint execution. It has vast advantages and provides a feasible case for the lack of interpretability and environmental adaptability in traditional deep reinforcement learning.

\citet{ref106lyu2022knowledge} proposed a KeGNN (Knowledge-enhanced graph neural network) for introducing prior knowledge in the form of first-order logic by stacking knowledge enhancement layer symbolic logic on top of the graph neural network for accurate reasoning on noisy graph data. This method first uses a graph neural network to extract node features and graph structure from graph structure data, represents each node as a feature vector related to text content, node attributes, and other information, and uses the graph structure to transfer and aggregate the characteristics of neighbor nodes. Feature information. KeGNN uses fuzzy logic to convert the continuous actual value output of GNN into a form that logical formulas can process; that is, it maps the true and false values of Boolean logic to continuous values in the [0, 1] interval and inputs the actual value of the node category into knowledge enhancement layer, and then use prior knowledge to perform learnable weight adjustments on these predictions. The KeGNN model is end-to-end differentiable, which also means that the GNN parameters and the weights of the knowledge enhancement layer can be learned simultaneously through the standard backpropagation algorithm. Symbolic logic in KeGNN exists as a knowledge enhancement layer, including prior knowledge in the form of first-order logic formulas and logic formulas for unary predicates and binary predicates. The former represents the attributes of nodes, and the latter describes node characteristics and relationships between nodes. Compared with traditional GNN models, KeGNN can improve classification accuracy to a certain extent in multiple benchmark data sets, which illustrates the effectiveness of KeGNN in processing graph-structured data.

\subsubsection{Symbolic:Symbolic Representation and Structure}
This category includes four studies in which neural networks extract features from physical interactions with the 3D world, visual modal data, and symbolic representations of environmental states. They use symbolic logic to describe environmental states, rules, and action effects and then make inferences based on this knowledge and regulation. Among these studies, those belonging to the neural-symbol generation classification include \cite{ref118balloch2023neuro}; \cite{ref141saravanakumar2021hierarchical}belongs to neuro-symoblic Enhancement, while those belonging to the neural-symbol collaboration include \cite{ref119franklin2020structured,ref120zellers2021piglet}.
\begin{table}[ht]
	\centering 
	\caption{Take environment and situation awareness data as input and symbolic representation and structure as symbolic method} 
	\label{tab:neuro_symbolic} 
	\begin{tabular}{
			>{\centering\arraybackslash}m{4cm} 
			>{\centering\arraybackslash}m{9cm} 
		}
		\toprule 
		Method of Collaboration & Papers \\
		\midrule 
		Neuro-symbolic generation & \cite{ref118balloch2023neuro} \\
		Neuro-symbolic enhancement &  \cite{ref141saravanakumar2021hierarchical} \\
		Neural-symbolic collaboration & \cite{ref119franklin2020structured,ref120zellers2021piglet} \\
		\bottomrule 
	\end{tabular}
\end{table}

\citet{ref120zellers2021piglet} proposed a framework, PIGLeT (Physical Interaction as Grounding for Language Transformers), that can extract common physical sense knowledge. This method can learn objects' physical properties and actions' consequences through interaction with the 3D simulation environment, which includes the materials the object is made of and the consequences of actions applied to the object. PIGLeT first uses a neural network to extract features from the physical interaction process with the 3D simulation environment. These features include the physical properties of objects and the actions that can be applied, such as understanding the physical consequences of different actions, such as moving and throwing on various objects. Physical dynamics models are then used to predict the outcomes of actions on objects in symbolic representations, converted into natural language descriptions. PIGLeT employs symbolic representations of physical dynamics models to capture object state changes due to interactions and natural language descriptions of interactions and states. Experimental results show that PIGLeT's understanding of the dynamics of the physical world exceeds that of large language models based on pure text learning. These results indicates that combining interactive learning and symbolic logic in a simulated environment can improve the machine's understanding of physical common sense.

In addition, \citet{ref118balloch2023neuro} proposed WorldCloner, a neuro-symbolic framework that adapts to environmental novelty changes by integrating neural networks and symbolic logic. WorldCloner can leverage its symbolic world model to learn efficient symbolic representations before environment transitions, quickly detect novelty, and adapt to novelty in a single trial. Specifically, this method first uses a neural network to extract features from the visual input of the environment state, such as the agent's location, types of surrounding objects, colors, etc. It encodes the above information into high-dimensional feature vectors. These vectors provide the information necessary to update the symbolic world model. When a state transition in the environment is inconsistent with the existing rules, the above information will adjust or add new rules to reflect the environmental changes. At the same time, the symbolic world model provides "imaginary" training data for the neural network by simulating environmental transformations so that the strategy can be updated and optimized without directly interacting with the environment. Symbolic logic in WorldCloner formally embodies a symbolic world or rule model. The model consists of logical expressions such as "if...then...". These rules describe the state transitions in the environment in detail. Compared with traditional model-free reinforcement learning and state-of-the-art world model methods like Dreamer V2, WorldCloner significantly improves handling environmental novelty. The specific performance is that when dealing with different types of novelty, such as DoorKeyChange, LavaProof, and LavaHurts, WorldCloner shows better or at least equivalent adaptation efficiency. Especially in the LavaProof scenario, DreamerV2 fails to adapt to the novelty of the environment, while WorldCloner can effectively discover and take advantage of new environmental changes to adjust strategies.
\subsubsection{Symbolic:Mathematical and Numerical Operations}
This classification includes one study. \citet{ref121landajuela2021discovering} proposed a new method, DSP (Deep Symbolic Policy), to solve the control problem in deep reinforcement learning by directly searching the symbolic policy space. The DSP framework uses an autoregressive RNN to extract features of the environment's observation or state data from the reinforcement learning environment. These features contain essential information, such as the position and speed of objects that control the current state of the task. The process starts with an empty expression and goes up to a sequence of mathematical operators and state variables. Therefore, DSP's understanding of the environmental state is transformed into a symbolic control strategy. The mathematical expression representing the policy can calculate one or more actions based on the current observation of the environment, which also means that RNN can learn how to map the environment state to a mathematical expression and use it as a policy to control the environment. These mathematical expressions directly affect the selection of actions in the environment.

Hence, DSP uses risk-seeking policy gradients to optimize the parameters of the RNN based on the rewards obtained by these actions in the environment, thereby improving the generated symbolic policy and maximizing the performance of the generated policy. In addition, DSP proposes an "anchoring" algorithm that can handle multi-dimensional action spaces. It uses pre-trained neural network-based strategies as temporary strategies and realizes the conversion from neural network strategies to symbolic strategies by gradually replacing them with pure symbolic strategies. DSP was tested in eight environments, including single-action and multi-action spaces, with benchmark environments performing continuous control tasks. The results showed that the symbolic policies discovered by DSP surpassed multiple state-of-the-art in terms of average ranking and average normalized plot reward, which indicates that this strategy generation method can produce a control strategy that is both efficient and easy to understand.
\subsection{Numerical Types and Mathematical Expressions}
This category includes 27 research results, all of which use neural networks to extract features from numerical data, sequence data, image data, and sensor data and then use mathematical expressions, mathematical equations, logical rules, constraints, probability models, and other symbolic logic to improve performance or interpretability. These studies can devide into three sub-categories by logical symbolic methods: logical rules and programming, symbolic representation and structure, and mathematics and numerical operations.
\subsubsection{Symbolic:Logic Rules and Programming}
This classification includes ten studies in which neural networks extract features from numerical data. At the same time, symbolic logic exists in the form of rules and constraints, propositional logic, ontology and reasoning mechanisms, and knowledge models. Research belonging to the neural-symbol generation category include \cite{ref122majumdar2023symbolic}; those belonging to the symbol-neural enhancement category include \cite{ref123machot2023bridging,ref124wang2018formal}; those belonging to the neural-symbol collaboration includes \cite{ref126long2019pde,ref127segler2018planning,ref128hooshyar2023augmenting,ref129thomas2022neuro,ref130amado2023robust,ref131daggitt2024vehicle,ref132ahmed2022semantic}.
\begin{table}[ht]
	\centering 
	\caption{Take numerical types and mathematical expressions as input and logic rules and programming as symbolic method} 
	\label{tab:neuro_symbolic} 
	\begin{tabular}{
			>{\centering\arraybackslash}m{4cm} 
			>{\centering\arraybackslash}m{9cm} 
		}
		\toprule 
		Method of Collaboration & Papers \\
		\midrule 
		Neuro-symbolic generation & \cite{ref122majumdar2023symbolic} \\
		Symbolic-neural enhancement & \cite{ref123machot2023bridging,ref124wang2018formal}\\
		Neural-symbolic collaboration & \cite{ref126long2019pde,ref127segler2018planning,ref128hooshyar2023augmenting,ref129thomas2022neuro,ref130amado2023robust,ref131daggitt2024vehicle,ref132ahmed2022semantic} \\
		\bottomrule 
	\end{tabular}
\end{table}

Among them, \citet{ref126long2019pde} proposed a method that can discover partial differential equations from observed dynamic data and predict the long-term dynamic behavior of these data in a noisy environment. This method first extracts features from the observation data of physical systems, such as fluid velocity fields or temperature distributions that change over time through convolution operations to approximate differential operators. The convolution kernel can be approximated by gradient, divergence, and Laplacian operators. It allows the neural network to learn the best approximation of these differential operations from the observation data and capture its spatial variation characteristics. In addition, PDEs(Partial differential equations) are also discretized through the forward Euler method in time and the finite difference method in space. This process can extract numerical information from the continuous physical process that the neural network can process to be regarded as a feature extraction process. Next, these approximations are input into SymNet(Symbolic Neural Network) as features and converted into symbolic logic. Symnet learns and approximates the PDE's nonlinear response function, revealing the PDE model's structure and form, equivalent to what will be learned from the data. Numerical characteristics are converted into symbolic mathematical descriptions of physical processes. Symbolic logic in PDE-Net 2.0 mainly exists in the form of SymNet. SymNet describes the nonlinear relationship of the system's dynamic behavior, including the approximation of nonlinear response functions and applying logic rules and constraints. The former learns nonlinear relationships in PDE through SymNet, and the latter integrates physical rules and mathematical constraints into the network learning process by imposing appropriate constraints on the convolution kernel and SymNet parameters. This method was tested using Burgers' diffusion and reaction-convection-diffusion equations. The results show that PDE-Net 2.0 can accurately restore the form of Burgers' equation, including the accurate coefficients of convection and diffusion terms, and restore the heat equation, including diffusion. The precise form, including the coefficients and the main structure of the Reaction-Convection-Diffusion Equation, including the coefficients of the reaction term, convection term, and diffusion term, are recovered from the data. The result shows that PDE-Net 2.0 can not only learn PDE with fixed coefficients but also handle the changes of parameters over time and space. This method can predict system behavior and reveal the underlying physical and mathematical mechanisms.

\citet{ref127segler2018planning} proposed a new method using CASP (computer-aided synthesis planning) to help chemists find better synthetic routes faster, 3N-MCTS. The authors used deep neural networks to learn reaction patterns and transformations in chemical reaction databases. The rules are then passed through three different neural networks to propose possible chemical transformations, predict reaction feasibility, and for sample transformations in the simulation phase. Specifically, the neural network is based on the molecular structure of reactants and products, using Extended-Connectivity Fingerprints like ECFP4 to represent molecules to extract features, including structural information and chemical transformation rules of chemical reactions from chemical reaction data. Symbolized chemical transformation rules automatically extracted from chemical reaction data are then used to predict whether a specific chemical transformation is likely to succeed. The process uses an expanded policy network to guide the search direction and propose possible chemical transformations during the search tree expansion phase, a feasibility prediction network to predict the feasibility of reactions proposed by the expanded policy network in natural chemical environments, and a rolling policy network to predict the feasibility of reactions proposed by the expanded policy network in simulations. The value of the synthetic position is estimated by sampling transformation in this stage. 3N-MCTS can find quicker synthesis paths than traditional computer-aided synthesis planning methods. In a double-blind AB test, the chemists participating in the evaluation were unable to significantly distinguish the quality difference between the synthetic pathways generated by 3N-MCTS and those reported in the literature, which means that the pathways generated by the Neuro-Symbolic AI method are qualitatively comparable to those of human experts. Designed paths are comparable.
\subsubsection{Symbolic:Symbolic Representation and Structure}
This combination includes two studies \cite{ref193biggio2021neural,ref133hasija2023neuro}. The former focuses on the neuro-symbolic generation, and the latter studies symbolic-neural enhancement, in which the neural network extracts features from the code or numerical input-output pairs of programming languages and uses symbolic logic methods such as abstract grammar tree or symbolic equation generation to represent high-level semantic representations.
\cite{ref133hasija2023neuro} proposed a new method for finding semantically similar code fragments in COBOL code. This approach defines a meta-model and instantiates it as an abstract syntax tree common between C and COBOL code as an intermediate representation that can capture the structure and logic of the code and serve as the symbolic logical form of the code. Using a neural network, this intermediate representation is extracted from the two programming language codes of C and COBOL. Then, the intermediate representation is converted into a one-dimensional serialized form using the traversal method. Finally, training and fine-tuning are performed on these linearized intermediate representations based on neural network models such as UnixCoder to learn the semantic similarities between code fragments. Symbolic logic exists in two primary forms in this method: intermediate representation and linearized intermediate representation. As a high-level abstraction of the code, the former embodies the program's logical structure and ignores specific grammatical details. At the same time, the latter enables the neural network to pass. This form learns the structure and semantics of code. The experiment verified the effectiveness of the code clone detection task on the COBOL test set by comparing random models, UniXCoder models fine-tuned for specific tasks, pre-trained UniXCoder models, and UniXCoder models fine-tuned with original C code. The UniXCoder model achieved a 36.36\% improvement in the MAP@2 indicator after being fine-tuned with SBT(Structure Based Traversal) IR(Intermediate Representation) of C code. At the same time, compared with fine-tuning with the original C code, the UniXCoder model fine-tuned with SBT IR of C code can migrate better. To COBOL code, zero-shot learning for cross-language code understanding is achieved.
\subsubsection{Symbolic:Mathematical and Numerical Operations}
The portfolio includes 15 studies in which neural networks extract features from experimental data, simulated data, time series signals, images, or numerical inputs in specific problem areas, such as structural engineering, physical science, chemistry, etc., then apply mathematical expressions, equations, or symbolic logic methods in the form of probability models. Mathematical derivation can transform features learned by neural networks into easily understood and explained forms, improving the model's ability to understand and predict data. Among these studies, those belonging to the neural-symbol generation classification include \cite{ref134petersen2019deep,ref135lample2019deep,ref136kubalik2023toward,ref137bendinelli2023controllable,ref138d2022deep,ref139bahmani2024discovering}; studies belonging to the symbolic-neural enhancement classification include \cite{ref140jia2023symbolic,ref141saravanakumar2021hierarchical,ref142mnih2014neural}; studies belonging to the neural-symbolic collaboration include \cite{ref143arabshahi2018combining,ref144mundhenk2021symbolic,ref145dutta2023s,ref146podina2022pinn,ref147trapiello2023verification,ref193biggio2021neural}.
\begin{table}[ht]
	\centering 
	\caption{Take numerical types and mathematical expressions as input and mathematical and numerical Operations as symbolic method} 
	\label{tab:neuro_symbolic}
	\begin{tabular}{
			>{\centering\arraybackslash}m{4cm} 
			>{\centering\arraybackslash}m{9cm} 
		}
		\toprule 
		Method of Collaboration & Papers \\
		\midrule 
		Neuro-symbolic generation & \cite{ref134petersen2019deep,ref135lample2019deep,ref136kubalik2023toward,ref137bendinelli2023controllable,ref138d2022deep,ref139bahmani2024discovering} \\
		Symbolic-neural enhancement & \cite{ref140jia2023symbolic,ref141saravanakumar2021hierarchical,ref142mnih2014neural}\\
		Neural-symbolic collaboration & \cite{ref143arabshahi2018combining,ref144mundhenk2021symbolic,ref145dutta2023s,ref146podina2022pinn,ref147trapiello2023verification,ref193biggio2021neural} \\
		\bottomrule
	\end{tabular}
\end{table}

\citet{ref146podina2022pinn} proposed a neural symbolic method to reconstruct the solution of the entire ordinary or partial differential equation in the case of sparse data. This method uses neural networks to extract features from existing numerical data of ordinary or partial differential equations. Usually, these numerical data describe the changes in the system state over time and space, so when faced with unknown physical laws or equations, The neural network can learn the system's dynamic characteristics from this data. The method then converts the numerical representation learned by the neural network into a symbolic equation through symbolic regression techniques such as AI Feynman. Symbolic logic in this study exists in two primary forms: the known part and the unknown part of the differential equation, where the former is a mathematical representation of an a priori understanding of the system dynamics and is given in the form of known differential operators; The operator is learned and represented by another neural network and then converted into a symbolic expression through symbolic regression technology. This part represents the unknown operators in the differential equation that data learning needs to discover. Experimental results show that the method performs strongly in several test cases. First, in the Lotka-Volterra scenario, the system can obtain good model recovery by increasing the number of calculation points under both noise-free and noisy data conditions; in the apoptosis model scenario, the learned function interacts with the actual mean squared error between solution and accurately discover the hidden terms with the mean square error even using only two-time points (initial condition t=0 and later time t=0.5) of noisy data $5 \times 10^{-3}$, can also reconstruct the PDE solution, and accurately discover hidden objects with a MSE(mean square error) of $3 \times 10^{-4}$ and a mean square error of $2 \times 10^{-2}$ item. The above experimental results demonstrate the effectiveness of this method in discovering and understanding hidden dynamic behaviors in complex systems.

\citet{ref140jia2023symbolic} proposed a SRNN (symbol-based recurrent neural network) that can model and predict the nonlinear response of concrete structures under seismic excitation without requiring large amounts of training data. SRNN uses neural networks to extract modal features such as displacement, velocity, and acceleration from the time history analysis of the structure's dynamic response and learn knowledge of the nonlinear dynamic model of the structure's behavior. Next, symbolic activation functions transform this nonlinear dynamic model into a set of ordinary differential equations that numerical integration methods can solve for engineers to understand and use easily. Symbolic logic, in the form of symbolic activation functions in this study, can discover mathematical expressions in the form of sine, cosine, square, and multiplication that describe the relationship between inputs and outputs. In addition, SRNN also uses hidden states to store nonlinear sequence information and provide the neural network with nonlinear characteristics of time series data. Experimental results show that SRNN has shown promising results in estimating the nonlinear response of structures. In the application case of a single-degree-of-freedom system, SRNN successfully learned the nonlinear behavior of the structural response and was able to predict the reaction under unseen ground motion accurately; for multi-degree-of-freedom systems, despite some challenges, SRNN can still better capture the nonlinear dynamic behavior of the structure, but the accuracy of the latter prediction drops slightly, with the correlation coefficient ($/rou$) varying between 0.83 and 0.88, which is somewhat lower than the performance of the single degree of freedom system.
\subsection{Structured Data}
This category includes 27 studies, all using neural networks to extract features from graph-structured, structured symbolic, and labeled parameter data. They then use symbolic logic, such as knowledge graphs, logical rules, parameter graphs, and label rules, to represent the structural and logical relationships between data. These studies apply three logical symbolic methods: logical rules and programming, symbolic representation and structure, and knowledge graphs and databases.
\subsubsection{Symbolic:Logic Rules and Programming}
The portfolio includes 15 studies in which neural networks extract features from structured symbolic, graph-structured, and time series data. On the other hand, symbolic logic utilizes directly defined logical rules, rule-based reasoning, or enhanced knowledge graph. Among these studies, those belonging to the neural-symbol generation classification include \cite{ref148singh2023neustip,ref149uria2023invariants,ref150finzel2022generating}; research belonging to the symbol-neural enhancement classification include \cite{ref151scassola2023conditioning,ref152cai2017making,ref153marra2021neural,ref154dong2019neural,ref155alshahrani2017neuro,ref156rivas2022neuro,ref157zhu2023approximate}; studies belonging to neural-symbolic collaboration include \cite{ref158sun2021neuro,ref159niu2021perform,ref160garg2020symbolic,ref161sen2021combining,ref162sen2022neuro}.
\begin{table}[ht]
	\centering
	\caption{Take structured data as input and mathematical and logic rules and programming as symbolic method} 
	\label{tab:neuro_symbolic}
	\begin{tabular}{
			>{\centering\arraybackslash}m{4cm} 
			>{\centering\arraybackslash}m{9cm} 
		}
		\toprule 
		Method of Collaboration & Papers \\
		\midrule 
		Neuro-symbolic generation & \cite{ref148singh2023neustip,ref149uria2023invariants,ref150finzel2022generating} \\
		Symbolic-neural enhancement & \cite{ref151scassola2023conditioning,ref152cai2017making,ref153marra2021neural,ref154dong2019neural,ref155alshahrani2017neuro,ref156rivas2022neuro,ref157zhu2023approximate}\\
		Neural-symbolic collaboration & \cite{ref158sun2021neuro,ref159niu2021perform,ref160garg2020symbolic,ref161sen2021combining,ref162sen2022neuro} \\
		\bottomrule 
	\end{tabular}
\end{table}

Among them, \citet{ref158sun2021neuro} proposed an NSPS (neuro-symbolic program search) method that improves the automation level of autonomous driving system design by automatically searching and synthesizing neural symbolic programs. The approach uses neural networks to extract features from structured, parametric observations as "attributes" in domain-specific languages representing streams of numerical data related to vehicle status and environment, such as waypoints, speed, acceleration, and bounding boxes. NSPS automatically searches a given neural symbol operation set and selects the necessary neural symbols to assemble into a program. The program is divided into two parts: a digital flow and a logical flow. The former processes sensory inputs such as vehicle speed and acceleration, and the latter executes rational judgment based on these inputs., such as whether the vehicle is in the deceleration stage approaching an intersection. At the same time, NSPS can query the target speed and waypoint index according to the current stage to implement corresponding vehicle operations. In this study, symbolic logic exists in the form of logical operations and numerical operations in a domain language designed for neural symbolic decision-making processes, where functions such as Intersect() and Union() perform numerical calculations based on number streams, and DecelerationPhase() Symbolic functions such as FollowUpPhase() and CatchUpPhase() perform logical judgments. Experimental results show that the NSDP(Neural-Symbolic Decision Program) obtained through the NSPS method achieves significant performance improvements in autonomous driving system design. NSDP can handle various driving scenarios, including car following, intersection merging, roundabout merging, and left turns at unseen intersections, and can achieve low collision rate, low acceleration, and low bump rate in various driving scenarios. The pure neural network approach produced smoother driving behavior.

\citet{ref148singh2023neustip} proposed a neural symbolic method, NeuSTIP (NeuroSymbolic Link and Time Interval Prediction), that simultaneously performs link and time interval predictions based on the temporal knowledge graph. This method innovatively introduces Allen time predicates that can ensure the temporal consistency of adjacent predicates in a given rule into rule learning and uses the learned rules to evaluate the confidence of candidate answers when performing link prediction and time interval prediction by designing a scoring function. NeuSTIP first uses neural networks to extract the relationships between entities from the quadruples (entity 1, relationship, entity 2, time interval) of Temporal Knowledge Graphs and the dynamic information of these relationships changing over time. NeuSTIP then uses Allen time predicates to learn temporal logic rules based on these features through neural networks. These rules are then used to reason and predict link prediction and time interval prediction tasks. For example, NeuSTIP can learn the rule: "If event A occurs in time interval T1, and event B occurs in time interval T2, and T1 and T2 satisfy a specific Allen time relationship, then event C can be predicted to occur in time interval T3." Experimental results show that the NeuSTIP model has achieved significant performance improvements in the temporal knowledge graph completion task. On the WIKIDATA12k data set, the NeuSTIP model's Mean Reciprocal Rank, Hits@1, and Hits@10 indicators all reached a high level; on the YAGO11k data set, the NeuSTIP model exceeded the TimePlex model and other benchmark models in all indicators. In addition, for the WIKIDATA12k and YAGO11k data sets, the NeuSTIP model surpassed the baseline HyTE, TNT-Complex, and Timeplex models in the aeIOU index, indicating that the temporal knowledge graph completion task can be effectively improved by learning and applying rules containing temporal logic—performance.

\subsubsection{Symbolic:Symbolic Representation and Structure}
This combination includes \cite{ref163cranmer2020discovering} and \cite{ref164riveret2020neuro}, where the former belongs to neural-symbol generation, and the latter belongs to symbolic-neural enhancement. The two neural networks abstract features from the dynamic data and labeled parameter data of the physical system and use explicit mathematical expressions in the form of parameter graphs and labels to guide the model learning process and enhance the interpretability of the model.
\begin{table}[ht]
	\centering 
	\caption{Take structured data as input and mathematical and symbolic representation and structure as symbolic method} 
	\label{tab:neuro_symbolic} 
	\begin{tabular}{
			>{\centering\arraybackslash}m{4cm} 
			>{\centering\arraybackslash}m{9cm} 
		}
		\toprule
		Method of Collaboration & Papers \\
		\midrule 
		Neuro-symbolic generation & \cite{ref163cranmer2020discovering} \\
		Symbolic-neural enhancement & \cite{ref164riveret2020neuro}\\
		\bottomrule
	\end{tabular}
\end{table}

\citet{ref164riveret2020neuro} proposed a new method that combines restricted Boltzmann machines and probabilistic semi-abstract argumentation to learn the probabilistic dependencies between argument labels by interpreting the argument labeling behind the data. This method uses trained restricted Boltzmann machines to extract relationships and patterns between argument labels from argument labeling data. Then, symbolic regression is used to convert the probabilistic dependencies the neural network learns into labels for argument diagrams. This transformation not only causes the network's output to contain predictions about the state of the argument but also provides explanations for these predictions. Symbolic logic in this study exists as markers in argument diagrams, representing interactions such as attack and support relationships between arguments and including the status of arguments such as acceptance, rejection, or pending. Experimental results show that compared with other standard machine learning technologies, NSAM(neuro-symbolic argumentation machine) shows advantages in handling probabilistic classification tasks. In experiments where swap noise is introduced, the performance of all different models decreases as the noise level increases. , NSAMs can mitigate the negative impact of noise through their built-in argumentation rules. Even when the noise level is high, the accuracy of NSAMs is still at least 25\% higher than other models. In addition to providing prediction results, NSAMs can also provide explanations of predictions by labeling argument diagrams.
\subsubsection{Symbolic:Knowledge Graphs and Databases}
The portfolio includes ten studies in which neural networks extract features from forms such as knowledge graphs, graph-structured data, or other symbolic logic data. In contrast, symbolic logic utilizes knowledge graphs, logical expressions, query structures, or rules to integrate domain knowledge, reasoning rules, or relationships. In these studies, those belonging to the neural-symbol generation classification include \cite{ref165ebrahimi2021neuro,ref166lemos2020neural}; research belonging to the symbol-neural enhancement classification include \cite{ref167raj2023neuro,ref168barcelo2023neuro,ref169carraro2023overcoming,ref170dold2022neuro,ref171chen2016multilingual,ref172cohen2020scalable}; studies belonging to neural-symbolic collaboration include\cite{ref173mota2016shared,ref192werner2023knowledge}.
\begin{table}[ht]
	\centering
	\caption{Take structured data as input and mathematical and knowledge graphs and databases as symbolic method}
	\label{tab:neuro_symbolic} 
	\begin{tabular}{
			>{\centering\arraybackslash}m{4cm}
			>{\centering\arraybackslash}m{9cm} 
		}
		\toprule
		Method of Collaboration & Papers \\
		\midrule 
		Neuro-symbolic generation & \cite{ref165ebrahimi2021neuro,ref166lemos2020neural} \\
		Symbolic-neural enhancement & \cite{ref167raj2023neuro,ref168barcelo2023neuro,ref169carraro2023overcoming,ref170dold2022neuro,ref171chen2016multilingual,ref172cohen2020scalable}\\
		Neural-symbolic collaboration & \cite{ref173mota2016shared,ref192werner2023knowledge} \\
		\bottomrule 
	\end{tabular}
\end{table}

\citet{ref171chen2016multilingual} proposed a new framework, MTransE (a translation-based model for multilingual knowledge graph embeddings), that achieves cross-language knowledge alignment by embedding multilingual knowledge graphs. First, MTransE learns its embedding vector representation in a low-dimensional space from the entities and relations of the knowledge graph. This step compresses the complex, high-dimensional knowledge graph information into a low-dimensional space convenient for calculation and alignment. The objective function enables the combination of entities and relationships in the knowledge graph to maintain the semantic relationships between them as much as possible in the embedding space. MTransE then uses Axis Calibration, Translation Vectors, and Linear Transformations to adjust and transform these embedding representations by minimizing the loss function of cross-language entity correspondences to achieve alignment between different language knowledge graphs. Among them, Axis Calibration makes entities and relationships with similar meanings closer in the embedding space of various languages by minimizing the distance between corresponding entities or relationship vectors across languages. Translation vectors can "translate" entity or relationship embedding vectors in one language to corresponding embedding vectors in another language, and linear transformations achieve cross-language knowledge alignment by learning a linear transformation matrix and mapping the embedding space of one language to the embedding space of another language. Experimental results show that some variants of MTransE, such as the linear transformation variants Var4 and Var5, are significantly better than other variants and baseline methods in cross-language entity matching tasks. The linear transformation technology is also validating a given cross-language ternary. Its effectiveness in maintaining semantic consistency between entities and relations has verified whether group pairs are correctly aligned. In addition, the MTransE model can be trained with only partial cross-language alignment of triples, can retain the critical properties of single-language embeddings while aligning cross-language knowledge, which means that it not only handles cross-language tasks but also effectively handles knowledge graph completion tasks within a single language.
\section{Multi-modal Non-Heterogeneous Neuro-symbolic AI}
This category includes 13 research results, all of which use neural networks to extract features from multiple modal data and then use symbolic logic such as knowledge graphs, logic programs, and symbolic rules to improve the system's inference and decision-making performance. These studies apply three logical symbolic methods: logical rules and programming, knowledge graphs and databases, and mathematics and numerical operations.
\begin{table}[ht]
	\centering 
	\caption{Articles belongs to multi-modal non-heterogeneous neuro-symbolic AI}
	\label{tab:neuro_symbolic} 
	\begin{tabular}{
			>{\centering\arraybackslash}m{3cm} 
			>{\centering\arraybackslash}m{6cm} 
			>{\centering\arraybackslash}m{5cm} 
		}
		\toprule 
		Papers & Input Modal & Logical Symbolic Form \\
		\midrule 
		\cite{ref174jiang2021lnn} & Text, Knowledge Graph & Rules defined by First-Order Logic\\
		\cite{ref175zheng2022jarvis} & Visual, Language & Task-Level and Action-Level Common Sense\\
		\cite{ref176chen2023genome} & Text, Image & Reasoning Steps generated by LLMs\\
		\cite{ref177tarau2021natlog} & Any modals that Python could process & Horn clauses and LD-resolution\\
		\cite{ref178zhang2018interpreting} & Financial Data, Logical Questions, Image & Linear Constraint\\
		\cite{ref179glanois2022neuro} & Text, Image & First-Order Logic\\
		\cite{ref180yi2018neural,ref195vedantam2019probabilistic} & Text, Image & Program Instructions\\
		\cite{ref182odense2022semantic} & Logical Expression, Graph & First-Order Logics\\
		\cite{ref183lazzari2024sandra} & Including but not limited to structured data, images, text, etc & Descriptions and Situations in Ontologys\\
		\cite{ref75marconato2023neuro} & sub-symbolic from images, text or other data forms & Prior Knowledge\\
		\cite{ref185siyaev2021neuro} & Voice, Text & Symbolic Program Executor\\
		\cite{ref186wang2019satnet} & Binary, probabilistic input & Differentiable (smoothed) Maximum Satisfiability (MAXSAT) Solver\\
		\bottomrule 
	\end{tabular}
\end{table}

\subsection{Symbolic:Logic Rules and Programming}
This category includes eight studies in which neural networks extract features from various model data, such as images and text, and then apply symbolic logic methods to improve the model's depth of understanding and reasoning performance. Among these studies, studies belonging to the neural-symbol generation classification include \cite{ref178zhang2018interpreting,ref179glanois2022neuro}; research belonging to the symbol-neural enhancement classification includes \cite{ref180yi2018neural}; studies belonging to neuro-symbolic collaboration include\cite{ref176chen2023genome,ref177tarau2021natlog,ref174jiang2021lnn,ref175zheng2022jarvis,ref182odense2022semantic}.
\begin{table}[ht]
	\centering 
	\caption{Symbolic method:Logic Rules and Programming} 
	\label{tab:neuro_symbolic} 
	\begin{tabular}{
			>{\centering\arraybackslash}m{4cm} 
			>{\centering\arraybackslash}m{9cm} 
		}
		\toprule 
		Method of Collaboration & Papers \\
		\midrule
		Neuro-symbolic generation & \cite{ref178zhang2018interpreting,ref179glanois2022neuro} \\
		Symbolic-neural enhancement & \cite{ref180yi2018neural}\\
		Neural-symbolic collaboration & \cite{ref176chen2023genome,ref177tarau2021natlog,ref174jiang2021lnn,ref175zheng2022jarvis,ref182odense2022semantic} \\
		\bottomrule 
	\end{tabular}
\end{table}

Among them, \citet{ref176chen2023genome} proposed a neural symbolic visual reasoning model, GENOME (GenerativE Neuro-symbOlic visual reasoning by growing and reusing ModulEs), that uses the programming capabilities of large language models (LLMs) to achieve modular translation of language descriptions. GENOME first uses large-scale language models to extract visual features such as objects and scenes and the relationship between these objects and scenes from images. LLMs also extract instructions or questions related to visual tasks, such as question analysis and keyword extraction from natural language texts. These two features are then subjected to logical operations through various modules and functions, such as using the object positioning module ``LOC'' to locate the position of specific objects in the image, using the counting module ``COUNT'' to count the number of objects that meet particular conditions, and using the conditional judgment module ``EVAL'' performs logical judgments based on specific attributes, etc. It is worth mentioning that LLMs generate these logic modules, and LLMs decide whether to create new modules based on the needs of the actual visual language task. Subsequently, the module execution phase performs reasoning on the input visual and language data by running the parsed symbolic logic operation sequence, combining the newly generated symbolic modules and modules in the existing module library, and finally generating the overall output of the task. Experiments show that the GENOME model performs well on standard Visual Question Answering and Referring Expression Comprehension tasks. In contrast, modules learned from one task can be seamlessly transferred to new tasks, and GENOME can also be trained with a small number of observation samples to adapt to new visual reasoning tasks. The above results show that GENOME can compete with existing models on standard visual reasoning tasks by generating and reusing modules and has excellent task adaptability and transfer learning capabilities.

\citet{ref177tarau2021natlog} proposed a lightweight logic programming language, a simple and practical Prolog-like language, Natlog, based on a unified execution model similar to Prolog. The syntax and semantics of this language are more simplified and can be tightly integrated into the Python-based deep learning ecosystem. In particular, Natlog can implement content-driven indexing based on ground-term data sets by rewriting the symbol indexing algorithm to delegate the same function to the neural network. Specifically, Natlog uses neural networks to process content-driven indexing of ground terms databases, learns patterns and associations from these multi-modal structured data, and performs content-driven indexing on input queries based on the patterns learned through training. This step is equivalent to using a neural network to provide an efficient retrieval mechanism to assist the symbolic logic engine in efficiently accessing and processing large-scale data sets. Subsequently, the relevant facts indexed by the neural network are sent to Natlog's logical reasoning engine, whose correctness is verified through unification and logical deduction steps, and the answer to the query is further deduced based on logical rules. The experimental section shows how to use logical queries to identify chemical elements with specific properties, use neural networks as content-driven indexers to predict database entries relevant to a given query and use these predictions in a logical inference process. The above experiments illustrate that Natlog can effectively retrieve and reason out query-related information from large-scale terrestrial terminology databases by integrating neural networks.
\subsection{Symbolic:knowledge Graphs and Databases}
This combination includes a total of three studies. \cite{ref75marconato2023neuro,ref183lazzari2024sandra,ref184siyaev2021neuro} all belong to Neuro-Symbolic collaborative classification. They apply knowledge graphs, ontology, logical rules, and other symbolic logic or structured knowledge to enhance the model's reasoning and explanation capabilities. They also provide the model with explicit understanding and prior knowledge about the world.

\citet{ref183lazzari2024sandra} proposed a neural symbolic reasoner, Sandra, that combines vector space representation and deductive reasoning, which enhances the model without significantly increasing computational complexity by mapping data into predefined symbolic descriptions. First, Sandra defined a set of descriptions and situations as symbolic logical forms. Description is an abstraction and generalization of a situation or phenomenon with multiple roles. These roles define how the elements in the description relate to each other. At the same time, a situation is a specific instance of a description represented by its corresponding position in the subspace defined by the Description. Each entity or attribute in the Situation is mapped to the vector corresponding to the roles of the Description. If the basis vectors in the subspace can linearly represent the situation vector in a description subspace, then we say the Situation satisfies the description. This method can process input data in multiple modalities such as text, images, and structured data and then map the input data to vectors in the vector space V defined by 'Sandra' through a neural network, where each Description is in the ontology. The corresponding vector subspace Vd is defined. By comparing the vector representation of the input data with the vector subspace of each Description, the system can infer which descriptions are consistent with the current input context and finally generate relevant outputs such as classification labels, reasoning explanations, etc., based on this explicit reasoning process. Experiments show that Sandra's performance under different configurations has been significantly improved. For example, in the '2x2' configuration, compared to the baseline model's 26.85\% accuracy, the Sandra model's accuracy is 45.75\%. In configuration C of the Fashion-MNIST (R-FMNIST) data set, after CNN is combined with Sandra, the accuracy rate increases from 43.13\% to 52.49\%. In addition, Lazzari et al. emphasized that the Sandra model is theoretically correct in line with the DnS model and can effectively provide interpretability and control over predefined vector spaces. This result reflects its ability to bridge vector and symbolic knowledge representation, improving the model's performance and enhancing its adaptability and interpretability in diverse data processing.
\subsection{Symbolic:Mathematical and Numerical Operations}
This classification includes one study. \citet{ref186wang2019satnet} solved the semi-definite program problem related to the MAXSAT (maximum satisfiability) problem by MAXSAT solver and using a fast coordinate descent method. This method is also called SATNet. SATNet first extracts features from numerical or logical data or image data. For logical data, logical encoding is used to represent the constraints of the problem, while for image data, a convolutional neural network is used to extract digital recognition features from Sudoku images. These features are then converted into a format suitable for logical reasoning—differentiable MAXSAT solvers. The direct logical data can be used as input for the MAXSAT problem.

In contrast, the image data must first be processed by a convolutional neural network to identify the numbers in the image, and the recognition results are converted into a logical format as the input of the MAXSAT problem. The MAXSAT solver then uses an optimization process to find a solution that satisfies all constraints and then converts the solution back into the representation of the original problem. At present, researchers have successfully used SATNet to learn logical structures and significantly improve performance in several tasks. SATNet can quickly help the model learn the objective function in the parity learning scenario and improve the test set on the test set within 20 cycles. The error rate converged to zero; in the Sudoku scenario, SATNet learned how to solve the standard 9×9 Sudoku puzzle, discovered and recovered the puzzle rules, and achieved 98.3\% accuracy on the test set, respectively. SATNet can effectively learn Sudoku game rules from image input for visual Sudoku tasks. It achieved a puzzle-solving accuracy of 63.2\% on the test set, close to the theoretical ``best'' test accuracy of 74.7\%. In this study, the MAXSAT solver is embedded in the learning process as a layer to integrate the processing capabilities of symbolic logic into the neural network architecture, thus belonging to the symbolic-neural enhancement classification.
\section{Single-modal Heterogeneous Neuro-symbolic AI}
\citet{ref187furlong2023modelling} proposed VSAs (Vector Symbolic Architectures) for simulating probability calculations and realizing symbolic logic and cognitive functions in brain model construction. In the VSAs framework, the features extracted from raw data by neural networks are converted into vector representations in high-dimensional space. Then, VSAs operate on the high-dimensional vectors to simulate symbolic logic. Specifically, it defines the Binding, Bundling, Similarity, and Unbinding operations, where the Binding operation uses the circular convolution or dot product of the vector to combine two vectors into a new vector that can uniquely represent the combination of the two original vectors; Bundling is the combination of multiple vectors are superposed together to form a new vector that roughly retains the original vector information; Similarity determines whether two symbols or concepts are similar or related by calculating the dot product or cosine similarity between two vectors; Unbinding is the inverse of binding operation used to extract a primitive vector from a bound vector. Based on these operations, VSA supports operations similar to traditional symbolic logic in high-dimensional vector spaces, such as building tree structures or graph structures representing complex data structures and relationships in vector space through binding and bundle operations. Alternatively, similarity calculations and unbundle perform pattern matching or rule application on vectors representing different concepts and rules to simulate the logical reasoning process. The logical reasoning part of the VSA architecture is transparent, but mapping raw data to high-dimensional vector space can still be regarded as a "black box" operation. Nevertheless, compared with traditional logical symbolic methods, the VSA architecture provides a parallel processing capability, which means that many logical symbolic operations can be processed simultaneously in vector space. This feature is significant for processing complex logical reasoning and large-scale knowledge bases.

\citet{ref188katz2021tunable} proposed a NVM (Neural Virtual Machine) for executing symbolic robot control algorithms. This method uses neural networks to perform symbolic operations by simulating the execution of a Turing-complete symbolic virtual machine. First, it extracts features from symbolic logic data through neural networks, converts symbolic logic operations into activity patterns and connection weights within the neural network, and uses the specific activation patterns of a group of neurons to represent variable names, operators, etc., in the program symbol. These activation patterns are predefined so the neural network can accurately represent and distinguish various program symbols. Then, specific layers and activity patterns of neural networks are used to describe the state of registers, memory, instruction pointers, etc., in a Turing-complete virtual machine, and state changes are represented by updating the corresponding neural activity. This way, symbolic operations such as arithmetic operations, logical judgments, conditional branches, and loops can be performed through predefined neural network patterns and dynamic weight adjustment. The compiled program can then be sent to NVM for processing as a series of instruction sequences. In addition, through specially designed neural network layers, symbolic decisions can be converted into executable control signals, such as motor commands or action sequences. An essential advantage of NVM is the ability to program and perform complex tasks using virtually any program logic, which is critical for robot development and operation.
\section{Multi-modal Heterogeneous Neuro-symbolic AI}
\citet{ref188katz2021tunable} proposed an LNN (Logical Neural Network) framework that integrates neural networks and logical symbol processing functions. The innovation of LNN is that neural networks and symbolic logic operate the same type of data in the same representation space. This framework avoids the use of additional middle layers to convert data types. Specifically, LNN supports using neural networks to extract features from raw data in multiple modalities, such as numerical, text, image, and sound data. More importantly, LNN corresponds logical symbols in propositions, predicates, etc., to one or a group of neurons, which means that the activation state of each neuron or neuron group represents the truth value state of the logical proposition, such as activation. The state indicates that the proposition is accurate, and the inactive state suggests that the proposition is false. At the same time, logical operations such as AND, OR, and NOT can also be implemented through specific activation functions and network structure design. For example, the AND operation can be constructed through the weighted sum of multiple inputs and a threshold activation function. The output neuron is activated only when all inputs are activated, which means the AND operation result is valid. The OR operation activates the output neuron when either input is valid, indicating that the result of this operation is valid. The NOT operation activates the output neuron when the input is inactive. This way, LNN can construct more complex logical expressions and support various logical symbol forms such as propositional, predicate, fuzzy, description, and temporal logic.

LNN adopts an end-to-end training method and does not require manual setting of rules or logical reasoning steps. It performs logical operations based on the learned parameters. Each network forward propagation is equivalent to performing a parameterized logical operation. For example, when we train an LNN that performs an AND operation, we can use the truth value status of two propositions as input and the AND operation result as the output. The training data set contains all possible truth value input combinations and the corresponding AND operation results. The network learns parameters through training to perform its AND operations accurately when receiving propositional states.

In traditional deep learning models, the internal hidden layers of the model are often challenging to interpret and treated as black boxes. It is difficult to accurately explain the specific meaning and role of each parameter, such as the weight and bias of the neuron, and how they work together to achieve the logical operation of the entire network. LNN attempts to use logic gates and map logic rules directly into the network's structure, but its learning process is still a ``black box.'' Although the method can perform specific logical operations, the detailed mechanism of how these logical operations are represented and processed inside the LNN needs to be more intuitive.

Despite this, LNN is still a meaningful attempt to use the same representation method in the same representation space to perform the operations of neural networks and logical symbols. First of all, it abandons the traditional representation conversion layer and attempts to adopt a fusion approach to process these different types of data and perform logical operations, which can achieve knowledge alignment of neural networks and symbolic logic more naturally and at the same time, complex data conversion and information loss are also avoided. In addition, since LNN directly maps logical operations into the neural network, the activation state of each neuron or neuron group can directly correspond to the truth value state of the logical proposition, so the decision-making process of LNN performing logical operations is more straightforward to explain. More importantly, this integrated processing method may bring new inspiration to the design of large language models, helping LLM stabilize internal concept representation and provide more accurate and interpretable logical chain reasoning capabilities.

In addition, the HDC (hyperdimensional computing) or VSA (vector symbolic architecture) method proposed by \cite{ref190kleyko2022survey,ref191kleyko2023survey} provides a different process from traditional neural networks and symbolic logic reasoning to implement Neuro-Symbolic AI. Chapter VI mentions that in this approach, data and concepts are represented as highly high-dimensional vectors capable of capturing complex patterns and relationships as a unified representation of symbolic and non-symbolic information. Therefore, traditional symbolic logic operations can be simulated by performing arithmetic and logical operations on high-dimensional vectors. At the same time, with the help of the orthogonality of vectors in high-dimensional space, HDC can retrieve stored information through simple approximate matching and support fast retrieval and associative memory. In addition, HDC methods can extract and generalize patterns from data and support complex decision-making and reasoning tasks by learning high-dimensional vector spaces, thus providing a natural and effective way to integrate symbolic logic and neural network processing. Because the above two reviews have introduced the HDC or VSA methods in detail, this article will not go into detail here.
\section{Dynamic Adaptive Neuro-symbolic AI}
Compared with multimodal heterogeneous neuro-symbolic AI, this classification can dynamically adjust and adapt to computing tasks regarding multimodal data processing, symbolic logic processing, and internal representation adjustment. Currently, no research meets the requirements of dynamic adaptive neuro-symbolic AI. Specifically, this classification is characterized by the following features.
\subsection{Automatic Selection and Integration of Appropriate Modal Data Processing Strategies}
First, such a system can automatically select and integrate the most suitable modal data processing strategy according to the needs and context of the specific task when performing feature extraction through neural networks. For example, when faced with a dual-modal task of visual and text, the system may prioritize using visual features for preliminary extraction. When the visual information is insufficient to support symbolic logic decision-making, it may combine the contextual information provided by the text modality for in-depth reasoning. The selection and integration of this strategy is not statically preset but dynamically generated through the system's learning and adjustment process. This capability also gives the dynamic adaptive neuro-symbolic AI system more efficient, more accurate, and more energy-friendly features when processing multi-modal data.
\subsection{Dynamically Adjust the Way Symbolic Logic is Processed}
Secondly, the dynamic adaptive neuro-symbolic AI system can automatically select the form of symbolic logic processing according to the task's requirements. This feature means the system can handle various logical reasoning tasks and dynamically select the most appropriate symbolic logic processing method based on different task characteristics. For example, when processing tasks that require complex logical reasoning, the system may adopt more sophisticated and complex logic rules, while when processing simple or intuitive tasks, it may adopt a more direct logical processing strategy. This dynamic adjustment capability improves the system's flexibility in processing logical reasoning tasks and optimizes reasoning efficiency and energy consumption.
\subsection{Self-adjust Internal Representation Based on Feedback and Task Performance}
Finally, the ability to self-adjust internal representation based on feedback and task performance means that the system can automatically adjust and optimize internal data representation and processing logic based on actual task execution results and performance evaluations. Moreover, this self-adjustment includes not only fine-tuning of model parameters but also fundamental adjustments to model structure and processing strategies. For example, the system may find that the processing method of a specific modal data is not effective enough during the processing of the task, so it can automatically enhance its logical processing module by adjusting the processing strategy or switching to a more complex logical reasoning module when processing a specific task. This self-adjustment capability based on task performance allows the neuro-symbolic AI system to continuously adapt to various task requirements and environmental challenges.
\section{Acknowledgment}
I would like to thank my supervisor VS Sheng,whose expertise was invaluable in formulating the research questions and methodology. Your insightful feedback pushed me to sharpen my thinking and brought my work to a higher level.
\bibliographystyle{ACM-Reference-Format}
\bibliography{reference2}
\end{document}